\documentclass{article}

\usepackage{microtype}
\usepackage{graphicx}
\usepackage{booktabs} 

\usepackage[accepted]{icml2023}

\usepackage{amsmath,amsthm,verbatim,amssymb,amsfonts,amscd, graphicx}
\usepackage{graphics}
\usepackage{centernot}
\usepackage{authblk}
\usepackage{paralist}

\theoremstyle{plain}
\newtheorem{theorem}{Theorem}

\newtheorem{lemma}{Lemma}

\newcommand{\sgn}{\mathrm{sgn}}
\newcommand{\E}{\mathbb{E}}

\newcommand{\p}{{\partial}}

\newcommand{\update}[1]{{\color{orange}#1}}

\usepackage[colorlinks,linkcolor=blue,citecolor=blue]{hyperref}

\usepackage[bottom]{footmisc}
\usepackage{caption}
\usepackage{helvet}  
\usepackage{courier}  
\usepackage{url}  
\usepackage{graphicx}  
\usepackage{multirow}
\usepackage{amsthm}
\usepackage{color}
\usepackage{MnSymbol}
\usepackage{makecell}
\usepackage{arydshln}
\usepackage{amsmath}

\usepackage{caption} 
\usepackage{natbib}
\usepackage{subcaption}
\usepackage{float}
\usepackage{textcomp}

\usepackage{wrapfig}
\usepackage{algorithm}
\usepackage{algorithmic}
\usepackage{paralist}
\usepackage{csquotes}

\icmltitlerunning{Sparsity by Redundancy}

\begin{document}

\twocolumn[
\icmltitle{\textit{spred}: Solving $L_1$ Penalty with SGD}



\icmlsetsymbol{equal}{*}

\begin{icmlauthorlist}
\icmlauthor{Liu Ziyin}{equal,utokyo}
\icmlauthor{Zihao Wang}{equal,hkust}

\end{icmlauthorlist}

\icmlaffiliation{utokyo}{The University of Tokyo}
\icmlaffiliation{hkust}{HKUST}

\icmlcorrespondingauthor{Liu Ziyin}{liu.ziyin.p@gmail.com}
\icmlcorrespondingauthor{Zihao Wang}{zwanggc@cse.ust.hk}

\icmlkeywords{Machine Learning, ICML}

\vskip 0.3in
]



\printAffiliationsAndNotice{\icmlEqualContribution} 

\begin{abstract}
    We propose to minimize a generic differentiable objective with $L_1$ constraint using a simple reparametrization and straightforward stochastic gradient descent. Our proposal is the direct generalization of previous ideas that the $L_1$ penalty may be equivalent to a differentiable reparametrization with weight decay. We prove that the proposed method, \textit{spred}, is an exact differentiable solver of $L_1$ and that the reparametrization trick is completely ``benign" for a generic nonconvex function. Practically, we demonstrate the usefulness of the method in (1) training sparse neural networks to perform gene selection tasks, which involves finding relevant features in a very high dimensional space, and (2) neural network compression task, to which previous attempts at applying the $L_1$-penalty have been unsuccessful. Conceptually, our result bridges the gap between the sparsity in deep learning and conventional statistical learning.
\end{abstract}


\vspace{-0mm}
\section{Introduction}

In many problems, optimization of an objective function under an $L_1$ constraint is of fundamental importance \citep{santosa1986linear, tibshirani1996regression, donoho2006compressed, sun2015feature, candes2008enhancing}. The advantage of the $L_1$ penalized solution is that they are sparse and thus highly interpretable, and it could be of great use if we can broadly apply the $L_1$ penalty to general problems. However, $L_1$ has only seen limited use in the case of simple models such as linear regression, logistic regression, or dictionary learning, where effective optimization methods are known to exist. As soon as the model becomes as complicated as a neural network, it is unknown how to optimize an $L_1$ penalty.

In contrast, with complicated models like neural networks, gradient descent (GD) has been the favored method of optimization because of its scalability on large-scale problems and simplicity of implementation. However, gradient descent has yet to be shown to work well in solving the $L_1$ penalty because the $L_1$ penalty is not differentiable at zero, precisely where the model becomes sparse. In fact, there is a large gap between the conventional $L_1$ learning and deep learning literature. Many tasks, such as feature selection, that $L_1$-based methods work well cannot be tackled by deep learning, and achieving sparsity in deep learning is almost never based on $L_1$. This gap between conventional statistics and deep learning is perhaps because no method has been demonstrated to efficiently solve the $L_1$ penalized objectives in general nonlinear settings, not to mention incorporating such methods within the standard backpropagation-based neural network training pipelines. Thus, optimizing a general nonconvex objective with $L_1$ regularization remains an important open problem. 

\begin{figure*}[t!]
    \centering
    \includegraphics[width=0.82\linewidth]{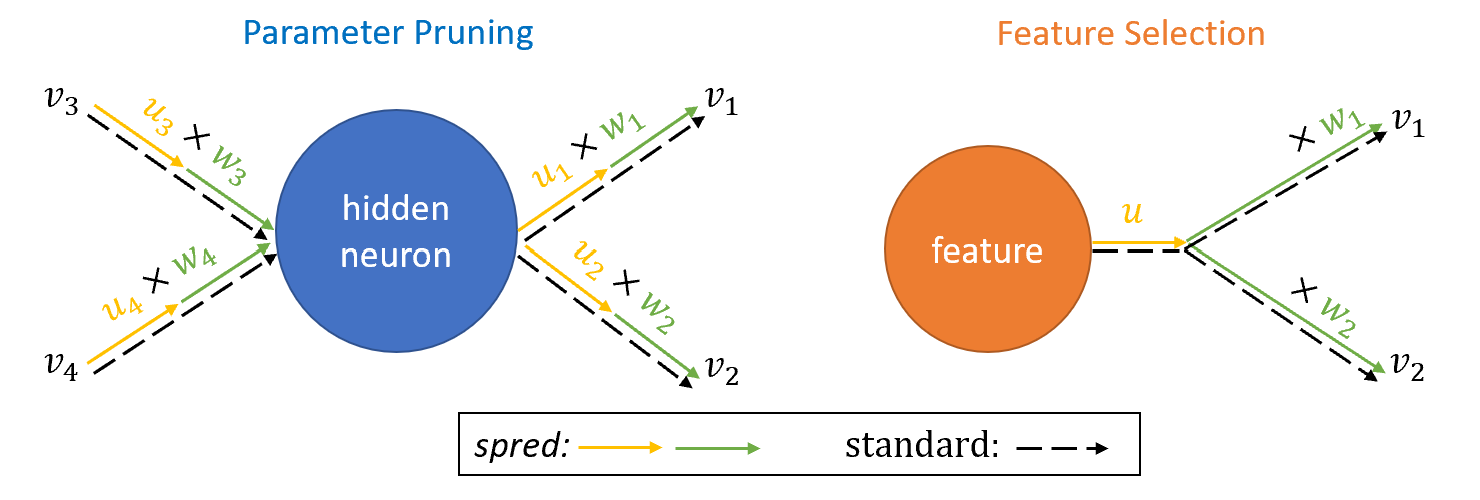}
    \vspace{-1em}
    \caption{\small Illustration of the \textit{spred} algorithm for achieving parameter sparsity (left) and feature selection (right). Essentially, the proposed algorithm creates redundant parameters and does not change the original architecture or training protocol. Therefore, the algorithm is compatible with pretraining.}\label{fig: method illustration}
    
    \includegraphics[width=0.27\linewidth]{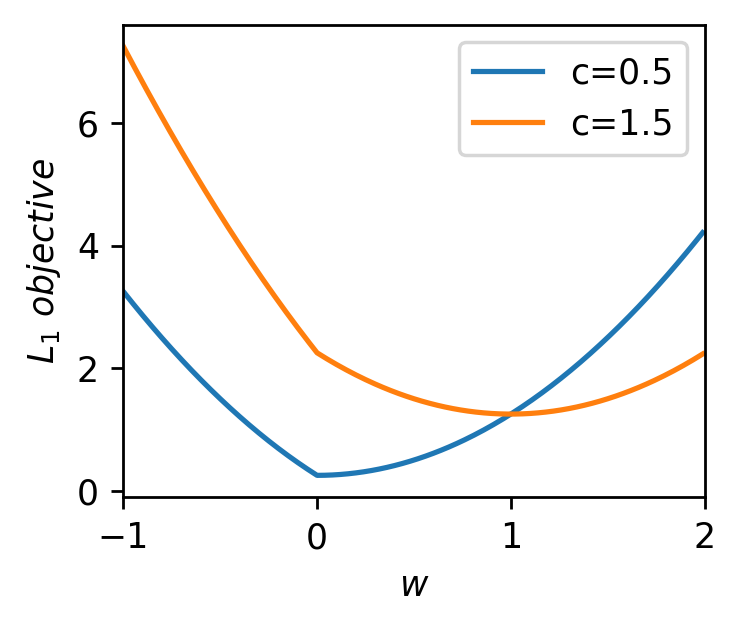}
    \includegraphics[width=0.28\linewidth]{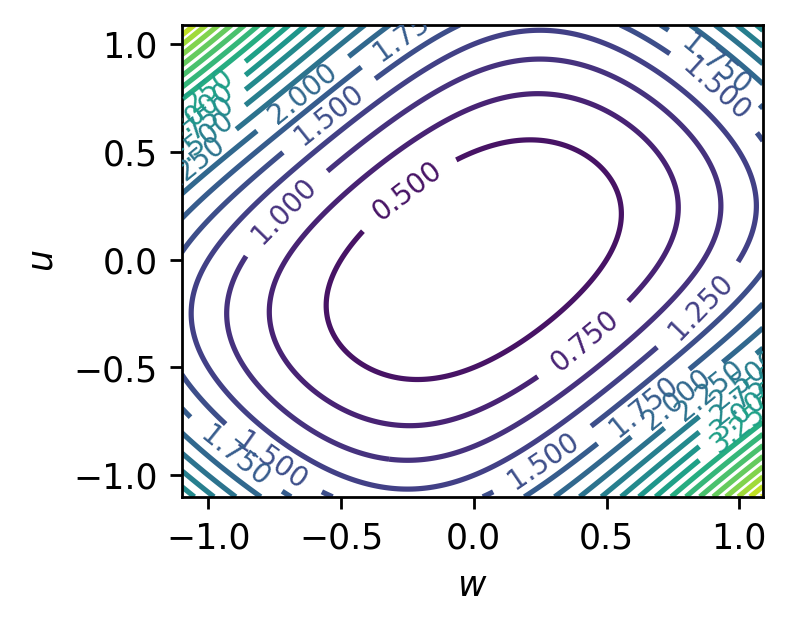}
    \includegraphics[width=0.28\linewidth]{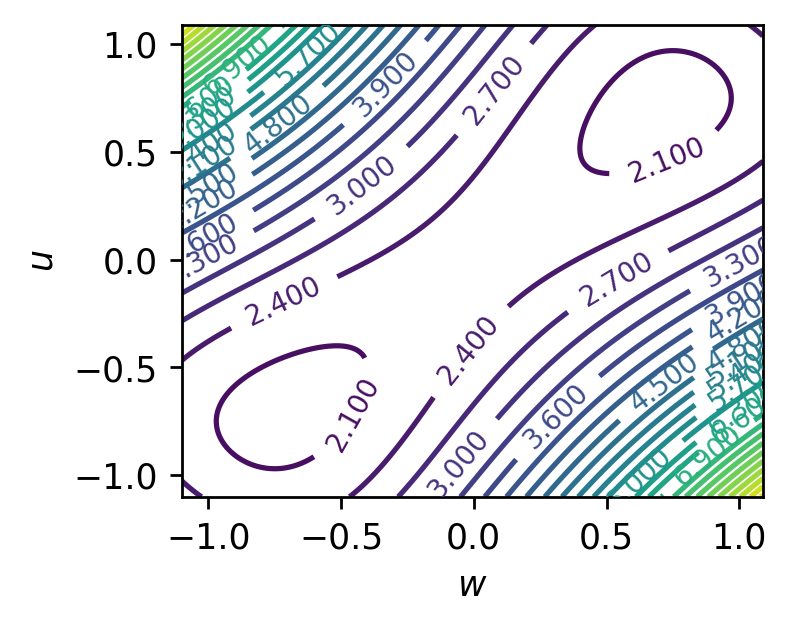}
    \vspace{-1em}
    \caption{\small 
    Loss landscape of the original $L_1$ regularized loss and the equivalent $L_2$ regularized redundant parametrization. With the redundant parametrization, the loss becomes smooth and differentiable. The reparametrization introduces one additional minimum but is entirely benign because the two minima are identical and converging to either achieves an equivalent performance. \textbf{Left}: the original 1d $L_1$ loss for $L_{L1}= (w-c)^2 + |w|$. \textbf{Mid}: reparametrized loss with $c=0.5$. \textbf{Right}: $c=1.5$. }
    \label{fig:landscape illustration}
    \vspace{-1em}

\end{figure*}

The foremost contribution of our work is to theoretically prove and empirically demonstrate that a reparametrization trick, also called the Hadamard parametrization, allows for solving arbitrary nonconvex objectives with $L_1$ regularization with gradient descent. The method is simple and takes only a few lines to implement in any modern deep-learning framework. Furthermore, we demonstrate that the proposed method is compatible with and can be boosted by common training tricks in deep learning, such as minibatch training, adaptive learning rates, and pretraining. See Figure~\ref{fig: method illustration} for an illustration.

\section{Related Works}

\textbf{L1 Penalty}. It is well-known that the $L_1$ penalty leads to a sparse solution \citep{wasserman2013all}. For linear models, the objectives with $L_1$ regularization are usually convex, but they are challenging to solve because the objective becomes non-differentiable precisely at the point where sparsity is achieved (namely, the origin). The mainstream literature often proposes special algorithms for solving the $L_1$ penalty for a specific task. For example, the original lasso paper suggests a method based on the quadratic programming algorithms \citep{tibshirani1996regression}. Later, algorithms such as coordinate descent \citep{friedman2010regularization} and least-angle regression (LARS) \citep{efron2004least} have been proposed as more efficient alternatives. The same problem also exists in the sparse multinomial logistic regression task \citep{cawley2006sparse}, which relies on a diagonal second-order coordinate descent algorithm. Another line of work proposes to use the iterative thresholding algorithms (ISTA) for solving lasso \citep{beck2009fast}, but it is unclear how ISTA-type algorithms could be generalized to solve general nonconvex problems. Instead of finding an efficient algorithm for a special $L_1$ problem, our strategy is to transform an $L_1$ problem into a differentiable problem for which the simplest gradient descent algorithms can be efficient.


\textbf{Redundant Parameterization}. The method we propose is based on a reparametrization trick of the $L_1$ loss function. The idea that a redundant parametrization with $L_2$ penalty has some resemblance to an $L_1$ penalty has a rather long history, and this resemblance has been utilized in various limited settings to solve an $L_1$ problem. \citet{grandvalet1998least} is one of the earliest to suggest an equivalence between $L_1$ and a redundant parametrization. However, this equivalence is only approximate. \citet{hoff2017lasso} theoretically studies the Hadamard parametrization in the context of generalized linear models and proposes to minimize the loss function by alternatively applying the solution of the ridge regression problem; notably, this work is the first to prove that not only the global minima of the redundant parametrization is equivalent to the $L_1$ global minima, but that all the local minima of the redundant parametrization are also local minima of the original $L_1$ objective, although only in case of linear models. \citet{poon2021smooth} studied the redundant parametrization in the case of a convex loss function and showed that all local minima of the redundant loss function are global and that the saddles are strict. In follow-up work, \citet{poon2022smooth} analyzed the optimization property of these convex loss functions. 

Compared to previous results, our result comprehensively characterizes all the saddle and local minima of the loss landscape of the redundant parametrization for a \textit{generic} and \textit{nonconvex} loss function. Our theoretical result, in turn, justifies the application of simple SGD to solve this problem and makes it possible to apply this method to highly complicated and practical problems, such as training a sparse neural network. Our motivation is also different from previous works. Previous works motivate the reparametrization trick from the viewpoint of solving the original Lasso problem, whereas our focus is on solving and understanding problems in deep learning. Application-wise, \citet{hoff2017lasso} applied the method to linear logistic regression.\cite{poon2021smooth} applied the method to lasso regression and optimal transport. In contrast, our work is also the first to identify and demonstrate its usage in contemporary deep learning.

\textbf{Sparsity in Deep Learning}. One important application of our theory is understanding and achieving any type of parameter sparsity in deep learning. There are two main reasons for introducing sparsity to the model. The first is that some level of sparsity often leads to better generalization performance; the second is that compressing the models can lead to more memory/computation-efficient deployment of the models \citep{gale2019state, blalock2020state}. However, none of the popular methods for sparsity in deep learning is based on the $L_1$ penalty, which is the favored method in conventional statistics. For example, pruning-based methods are the dominant strategies in deep learning \citep{lecun1989optimal}. However, such methods are not satisfactory from a principled perspective because the pruning part is separated from the training, and it is hard to understand what these pruning procedures are optimizing. 


\vspace{-0mm}
\section{Algorithm and Theory}
\vspace{-0mm}
In this section, we first introduce the reparametrization trick. We then present our theoretical results, which establish that the reparametrization trick does not make the landscape more complicated. All the proofs are presented in Appendix~\ref{app sec: proofs}.

\vspace{-0mm}
\subsection{Landscape of the Reparametrization Trick}
\vspace{-0mm}

Consider a generic objective function $L(V_s, V_d)$ that depends on two sets of learnable parameters $V_s$ and $V_d$, where the subscript $s$ stands for ``sparse," and $d$ stands for ``dense." Often, we want to find a sparse set of parameters $V_s$ that minimizes $L$. The conventional way to achieve this is by minimizing the loss function with an $L_1$ penalty of strength $2\kappa$:
\begin{equation}\label{eq: l1 loss general}
    \min_{V_s, V_d} L(V_s, V_d) +  2\kappa||V_s||_1.
\end{equation}
We will refer to $L(V_s, V_d) +  2\kappa||V_s||_1$ as $L_{L1}$. Under suitable conditions for $L$, the solutions of $L(V_s, V_d)$ will feature both (1) sparsity and (2) shrinkage of the norm of the solution $V_s$, and thus one can perform variable selection and overfitting avoidance at the same time. A primary obstacle that has prevented a scalable optimization of Eq.~\eqref{eq: l1 loss general} with gradient descent algorithms is that it is non-differentiable at the points where sparsity is achieved. The optimization problem only has efficient algorithms when the loss function belongs to a restrictive set of families. See Figure~\ref{fig:landscape illustration}.

Let $\odot$ denote the element-wise product. The following theorem derives a precise equivalence of Eq.~\eqref{eq: l1 loss general} with a redundantly parameterized objective. 
\begin{theorem}\label{theo: sparsity of parameters}
    Let $\alpha\beta = \kappa^2$ and
    \begin{equation}\label{eq: rs loss general}
        L_{sr}(U, W, V_d) := L(U\odot W, V_d) + \alpha ||U||^2 + \beta ||W||^2.
    \end{equation}
    Then, $(U,W, V_d)$ is a global minimum of Eq.~\eqref{eq: rs loss general} if and only if (a) $|U_i| = |W_i|$ for all $i$ and (b) $(U\odot W, V_d)$ is a global minimum of Eq.~\eqref{eq: l1 loss general}.\footnote{In this work, we use the letter $L$ exclusively for the part of loss function that does not contain $L_1$ or $L_2$ penalty.}
\end{theorem}
Because having $V_d$ in the loss function or not does not change the proof, we omit writing $V_d$ from this point on. We note that the suggestion that this reparametrization trick is equivalent to the $L_1$ penalty at global minima appeared in previous works under various restricted settings. A limited version of this theorem appeared in \citet{hoff2017lasso} in the context of a linear model. \citet{poon2021smooth} proved this equivalence in the global minimum when the landscape is convex. 

The subscript $sr$ stands for ``sparsity by redundancy." When $L$ is $n$-time differentiable, the objective $L_{sr}$ is also $n$-time differentiable. It is thus tempting to apply simple gradient-based optimization methods to optimize this alternative objective when $L$ itself is differentiable. When $L$ is twice-differentiable, one can also apply second-order methods for acceleration. As an example of $L$, consider the case when $L$ is a training-set-dependent loss function (such as in deep learning), and the parameters $V_s$ and $V_d$ are learnable weights of a nonlinear neural network. In this case, one can write $L_{sr}$ as
\begin{equation}
    \frac{1}{N}\sum_{i=1}^{N} \ell(f_{w}(x_i), y_i) + \alpha ||U||^2 + \beta ||W||^2,
\end{equation}
where $w=(U, W, V_d)$ denotes the total set of parameters we want to minimize, and $(x_i,y_i)$ are data point pairs of an empirical dataset. For a deep learning practitioner, it feels intuitive to solve this loss function with popular deep learning training methods. Additionally, $L_2$ regularization can be implemented efficiently as weight decay as in the standard deep learning frameworks. Section~\ref{sec: examples} provides several specific examples of this redundant parametrization.

However, the equivalence in the global minimum is insufficient to motivate an application of SGD to it because gradient descent is local, and if this parametrization induces many bad minima, SGD can still fail badly. An important question is thus whether this redundant parametrization has made the optimization process more difficult for SGD or not. We now show that it {does not}, in the sense that all local minima of Eq.~\eqref{eq: rs loss general} faithfully reproduce the local minima of the original loss and vice versa. Thus, the redundant parametrization cannot introduce new bad minima to the loss landscape.

\begin{theorem}\label{theo: benign landscape}
    All stationary points of Eq.~\eqref{eq: rs loss general} satisfy $|U_i| = |W_i|$. Additionally, $(U,W)$ is a local minimum of Eq.~\eqref{eq: rs loss general} if and only if (a) $V=U\odot W$ is a local minimum of Eq.~\eqref{eq: l1 loss general} and (b) $|U_i| = |W_i|$.
\end{theorem}
Namely, one can partition all of the local minima of $L_{rs}$ into exclusive and equivalent sets, such that these sets have a one-to-one mapping with the local minima in the corresponding $L_{L1}$. We are the first to prove this one-to-one mapping relation for a general loss function. This proposition thus offers a partial theoretical explanation to our empirical observation that optimizing Eq.~\eqref{eq: rs loss general} is no more difficult (and often much easier) than the original $L_1$-regularized loss. A corollary of this theorem reduces to the main theorem of \citet{poon2021smooth}, which states that if $L$ is convex (such as in Lasso), then every local minimum of $L_{rs}$ is global. A crucial new insight we offer is that one can still converge to a bad minimum for a general landscape, but this only happens because the original $L_{L1}$ has bad minima, not because of the reparametrization trick.

Still, this alone is insufficient to imply that GD can navigate this landscape easily because gradient descent can get stuck on saddle points easily \citep{du2017gradient, ziyin2021sgd}. In particular, GD often has a problem escaping higher-order saddle points where the Hessian eigenvalues along escaping directions vanish. The following theorem shows that this is also not a problem for the reparametrization trick because the strength of the gradient is as strong as the original $L_{L1}$.

    

\begin{theorem}\label{theo: benign saddle}
    Let $|U|= |W|$, $V=U\odot W$ and $L$ be everywhere differentiable. Then, for every infinitesimal variation $\delta V$,
    \begin{enumerate}
        \item if $L_{L1}(V)$ is directionally differentiable in $\delta V$, there exist variations $\delta W, \delta U\in \Theta(\delta V )$ such that $ L_{L1}(V+ \delta V)=  L_{rs}(U + \delta U, W+ \delta W)$;
        \item if $L_{L1}(V)$ is not directionally differentiable in $\delta V$, there exist variations $\delta W, \delta U\in \Theta\left((\delta V)^{0.5} \right)$ such that $ L_{L1}(V+ \delta V)=  L_{rs}(U + \delta U, W+ \delta W)$.
    \end{enumerate}
\end{theorem}
Namely, away from nondifferential points of $L_{L1}$, the reparametrized landscape is qualitatively the same as the original landscape, and escaping the saddles in the reparametrized landscape must be no harder than escaping the original saddle. If GD finds it difficult to escape a saddle point, it must be because the original $L_{L1}$ contains a difficult saddle. All nondifferentiable points of $L_{L1}$ occur at a sparse solution where some parameters are zero. Here, the first-order derivative is discontinuous, and the variation of the $L_{L1}$ is thus first-order in $\delta V$. This implies that the variation in the corresponding $L_{rs}$ is second-order in $\delta U$ and $\delta W$ and that the Hessian of $L_{rs}$ should have at least one negative eigenvalue, which implies that escaping from these points should be of no problem to gradient descent \citep{jin2017escape}. Combined, Theorem~\ref{theo: benign landscape} and \ref{theo: benign saddle} directly motivate the application of stochastic gradient descent to any problem that SGD has been demonstrated efficient for, an important example being a neural network. 

In more general scenarios, one is interested in a structured sparsity, where a group of parameters is encouraged to be sparse simultaneously. It suffices to consider the case when there is a single group because one can add $L_1$ penalty recursively to prove the general multigroup case:
\begin{equation}\label{eq: group L1}
    L(V_s, V_d) + \kappa |V_s|_2.
\end{equation}
The following theorem gives the equivalent redundant form.

\begin{theorem}\label{theo: main 2}
    Let $\alpha\beta = \kappa^2$ and
    \begin{equation}\label{eq: group rs loss}
        L_{sr}(u, W, V_d) := L(uW, V_d) + \alpha u^2 + \beta ||W||^2.
    \end{equation}
    Then, $(u,W, V_d)$ is a global minimum of Eq.~\eqref{eq: group rs loss} if and only if (a) $|u| = ||W||_2$ for all $i$ and (b) $(u W, V_d)$ is a global minimum of Eq.~\eqref{eq: group L1}.
\end{theorem}
Namely, every $L_1$ group only requires one additional parameter to sparsify. Note that recursively applying Theorem~\ref{theo: main 2} and setting $W$ to have dimension $1$ allows us to recover Theorem~\ref{theo: sparsity of parameters}.\footnote{{Note that when $L$ is a linear regression objective, the loss function is equivalent to the group lasso.}} The above theory justifies the application of the reparametrization trick to any sparsity-related tasks in deep learning. For completeness, we give an explicit algorithm in Algorithm~\ref{alg:spred general} and \ref{alg:spred structured}. Let $m$ be the number of groups. This algorithm adds $m$ parameters to the training process. Consequently, it has the same complexity as the standard deep learning training algorithms such as SGD because it, at most, doubles the memory and computation cost of training and does not incur additional costs for inference. For the ResNet18/CIFAR10 experiment we performed, each iteration of training with \textit{spred} takes less than $5\%$ more time than the standard training, much lower than the worst-case upper bound of $100\%$.


\begin{algorithm}[h]
\caption{\textit{spred} algorithm for parameter sparsity}\label{alg:spred general}
\begin{algorithmic}
   \STATE \textbf{Input}: loss function $L(V_s, V_d)$, parameter $V_s, V_d$, $L_1$ regularization strength $2\kappa$
   \STATE Initialize $W,\ U$
   \STATE Solve (with SGD, Adam, LBGFS, etc.) \newline $\min_{W, U, V_d} L(U \odot W, V_d) + \kappa (||W||_2^2 + ||U||^2 )$
   \STATE \textbf{Output}: $V^*= U\odot W$
\end{algorithmic}

\end{algorithm}
\begin{algorithm}[h]
\caption{\textit{spred} algorithm for structured sparsity}\label{alg:spred structured}
\begin{algorithmic}
   \STATE \textbf{Input}: loss function $L(V_s, V_d)$, parameter $V_s, V_d$, $L_1$ regularization strength $2\kappa$
   \STATE Initialize $W,\ u$
   \STATE Solve $\min_{W, u, V_d} L(u W, V_d) + \kappa (||W||_2^2 + u^2 )$
   \STATE \textbf{Output}: $V^*= uW$
\end{algorithmic}
\end{algorithm}

\textit{Implementation and practical remarks}. First, multiple ways exist to initialize the redundant parameters $W$ and $U$. One way is to initialize $W$ with, say, the Kaiming init., and $U$ to be of variance $1$. The other way is to give both variables the same variance by, e.g., taking the squared root of the standard initialization methods. A question is whether one should initialize with a balanced norm: $|u|=|w|$. Our initial experiments find no significant difference between making the norm balanced or not at initialization, and we recommend not balancing the weights as a default setting. Secondly, even if one only wants to add $L_1$ to one layer, one should also add a small weight decay to all the other layers to prevent the model from diverging. Lastly, while the proposed method does not require a threshold to reach a sparse solution, it could reduce the training time without affecting the performance by stopping earlier and pruning at a small threshold. Our experiments suggest that $10^{-6}$ is often a reasonable threshold for linear models and $10^{-3}$ for neural networks. 

\vspace{-0mm}
\subsection{Examples}\label{sec: examples}
\vspace{-0mm}

It is instructive to consider two examples to understand better how to apply the \textit{spred} parametrization.

\textbf{Example 1} ({lasso}). The lasso objective is $L(V_s) = \sum_i(V_s^T x_i -y_i)^2 + 2 \kappa ||V_s||_1$. The equivalent \textit{spred} loss is
\begin{equation}
    L(U, W) = \sum_i((U \odot W)^T x_i -y_i)^2 + \kappa (||W||^2 + ||U||^2),
\end{equation}
where $V_d$ is the empty set.

\textbf{Example 2} (unstructured sparsity in two-layer tanh nets). Let both the input and the label be one-dimensional. Also, let $V_s=(V_1, V_2)$ be the union of the first layer weight matrix $V_1$
 and the second layer weight $V_2$
. With the MSE objective, the original loss is 
$L(V_s)  = \sum_i(V_1  {\rm \tanh}(V_2 x_i) -y_i)^2 + 2 \kappa (||V_1||_1 + ||V_2||_1)$. The equivalent \textit{spred} loss is then
\begin{align}
    L(U, W) =& \sum_i((U_2 \odot W_2)^T {\rm \tanh} ((U_1 \odot W_1)x_i) -y_i)^2 \nonumber\\
    &+ \kappa (||W_1||^2 + ||W_2||^2 + ||U_1||^2 + ||U_2||^2),
\end{align}
where we have also partitioned the parameters into those of the two layers, respectively: $U=(U_1, U_2)$ and $U=(U_1, U_2)$. Here, $V_d$ is also the empty set.

See Figure~\ref{fig: method illustration} for an illustration. Also, see the next section for an example of structured sparsity.

\vspace{-0mm}
\section{Experiments}
\vspace{-0mm}

In this section, we empirically validate that \textit{spred} is useful for sparsity-related tasks in deep learning.\footnote{Code: \url{https://github.com/zihao-wang/spred}} We first demonstrate the correctness of the proposed approach for the classical lasso problem. Then, we apply the algorithm to two deep learning problems: (1) high dimensional nonlinear feature selection on gene datasets; (2) neural network compression.

\vspace{-0mm}
\subsection{Lasso}
\vspace{-0mm}

\begin{figure}
    \centering
    \includegraphics[trim={1mm 0 0 0},clip, width=0.95\linewidth]{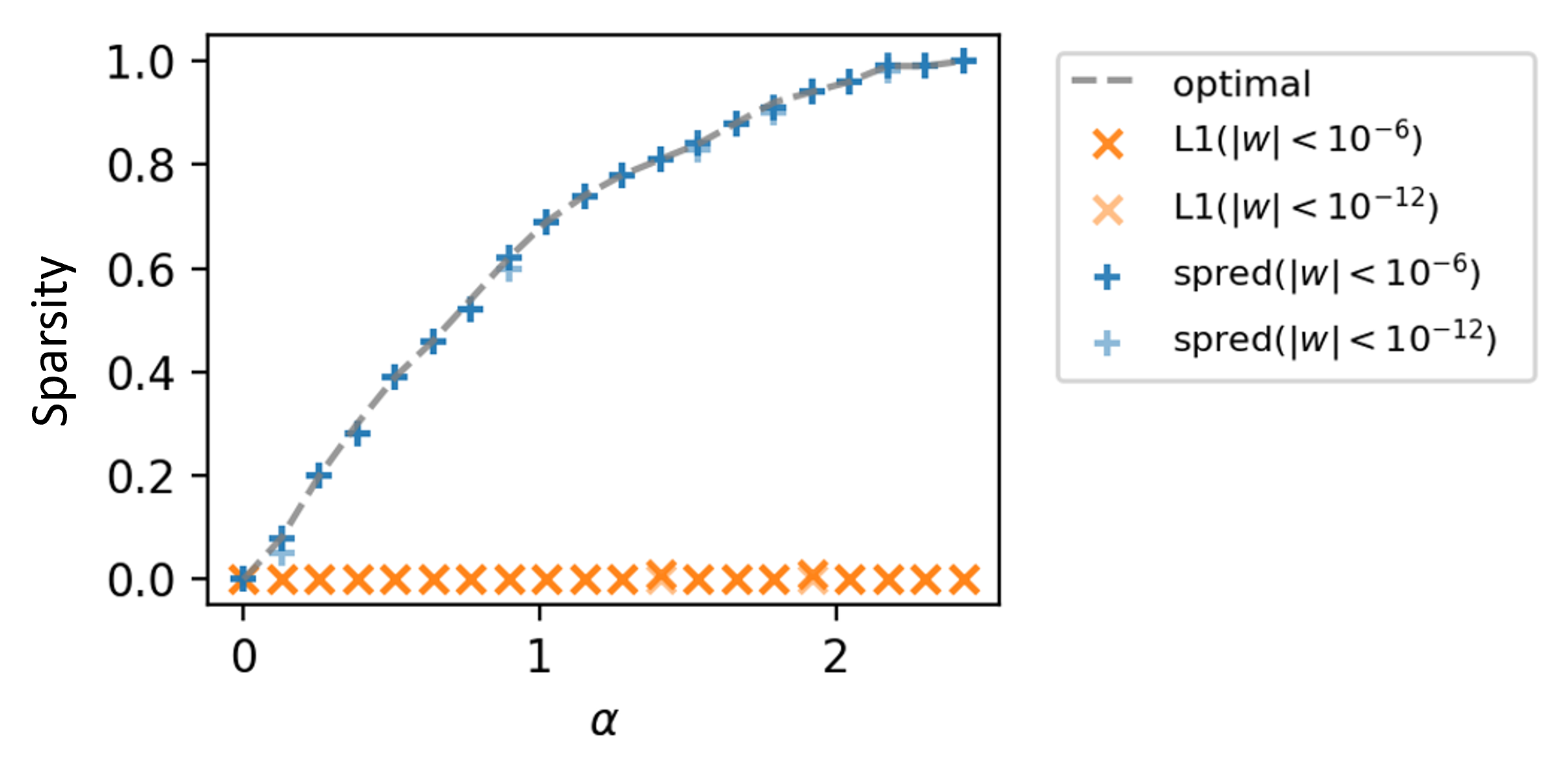}
    \vspace{-1em}
    \caption{\small \textit{spred} reaches the theoretical optimal solution when solving lasso across different values of $\alpha$. The dashed line shows the closed-form solution. \textit{L1}: $L_1$ regularized least square regression solved by gradient descent; \textit{spred}: the proposed method.} 
    \label{fig:lasso}
    \vspace{-1em}
\end{figure}

We illustrate the correctness of the proposed algorithm on the well-understood lasso problem. We first consider the case of an orthogonal input distribution. In this case, the closed-form solution for lasso is known, allowing us to evaluate whether the method can reach the optimal lasso solutions. For illustration, we also show the performance of the naive gradient-descent baseline: directly applying gradient descent to the original lasso objective, denoted as \textit{L1}. While one does not expect this method to work, it has been the popular way in deep learning to optimize the $L_1$ penalty (for example, see \citet{han2015deep} and \citet{scardapane2017group}). We choose both gradient descent and Adam optimizers to optimize \textit{spred}, as well as the original $L_1$ regularized mean square error objective. The learning rate is chosen from $\{1, 0.1, 0.01, 0.001\}$. The final result is chosen from the best setting. Figure~\ref{fig:lasso} shows that \textit{spred} agrees with the closed-form solution for all sparsity levels and for two different levels of accuracy, while the naive gradient-based method never reached a sparse solution. Our experiments also show that the convergence speed of \textit{spred} is similar to the standard $L_1$ optimization methods such as coordinate descent or LARS. See Appendix~\ref{app sec: exp}.

\begin{table*}[t]
\vspace{0em}
\caption{\small Prediction accuracy for the gene selection task for cancer diagnosis and survival time prediction. All tasks are classification tasks. On average, each dataset contains $300$ data points, each with $40000$ feature dimensions, and labeled into $10$ classes. See Table~\ref{tab: cancer dataset statistics} for more description.}
\vspace{-1em}
\label{tab: feature selection results}
\centering
\begin{tabular}{lrrrrr}
\hline
\multirow{2}{*}{Dataset}  & \multirow{2}{*}{HSICLasso} & \multirow{2}{*}{MLP (WD)} & \multirow{2}{*}{MLP (L1)} & \multicolumn{2}{c}{\textit{spred}}        \\ \cline{5-6} 
                              &                            &                           &                          & $f_l$ only          & $f_l$ and $f_n$     \\ \hline
GDS1815                           & $11.62\pm0.29$             & $0.56 \pm 0.22$           & $7.75 \pm 0.55$          & $17.75\pm0.77$      & $\bf{19.31\pm0.70}$ \\
GDS1816                              & $13.68\pm0.06$             & $0.31\pm 0.13$            & $7.12 \pm 0.75$          & $17.43\pm0.79$      & $\bf{18.75\pm0.77}$ \\
GDS3268                              & $\bf{30.69\pm0.44}$        & $3.03\pm 0.41$            & $15.90 \pm 0.81$         & $25.90\pm0.59$      & $27.86\pm0.65$      \\
GDS3952                               & $\bf{45.61\pm0.52}$        & $14.92\pm 1.14$           & $17.92 \pm 0.92$         & $37.00\pm1.22$      & $\bf{46.76\pm1.55}$ \\
GDS4761                                  & $42.63\pm0.51$             & $50.79\pm 2.48$           & $12.62 \pm 1.78$         & $\bf{60.26\pm2.37}$ & $\bf{57.63\pm2.09}$ \\
GDS5027                                 & $23.51\pm0.10$             & $2.55\pm 0.48$            &            $15.47\pm 0.98$              & $\bf{30.37\pm0.97}$ & $\bf{30.95\pm0.94}$ \\ \hline
\end{tabular}
\vspace{-1em}
\end{table*}

\vspace{-0mm}
\subsection{Nonlinear Feature Selection}
\vspace{-0mm}
The common gene selection tasks have a feature dimension of order $10^4-10^5$ (the size of the human genome), and the number of samples (often the number of patients) is of order $10^2$  \citep{shevade2003simple, sun2015feature}. These tasks can be seen as a "transpose" of MNIST and are the direct opposite of the tasks that deep learning is good at. Additionally, one indispensable part of these tasks is that we want to not only make generalizable predictions but also pinpoint the relevant genes that have a direct physiological consequence. For example, out of roughly $50000$ genes of human beings, we want to know which gene is the closest associated with, say, hemophilia -- such a requirement for interpretability is also challenging for deep learning. At the heart of this problem is a feature selection problem. Existing feature-selection methods based on $L_1$ penalty are predominantly linear. The nonlinear methods are often kernel-based, where the nonlinearity comes from an unlearnable kernel. While neural networks have fantastic capabilities in capturing nonlinear associations in the data, it is generally unknown how to apply deep learning to this problem.

In this section, we demonstrate how \textit{spred} offers a direct way to apply deep learning nonlinear feature selection. To the best of our knowledge, no deep learning method has been shown to work for these tasks (for a review, see \citet{montesinos2021review}). We compare with relevant baselines on $6$ public cancer classification datasets based on microarray gene expression features from the Gene Expression Omnibus, including two datasets on glioma (\#1815, \#1816), three on breast cancer (\#3952, \#4761, \#5027), and one on ulcerative colitis (\#3268). More detailed descriptions of the datasets are in the appendix. 

At the same time, linear models have been found to work reasonably well for these tasks. Thus, one would like to make feature selections based on both linear and nonlinear models. The proposed method allows one to achieve this goal easily: we demonstrate how to perform feature selection with an ensemble of models using \textit{spred}. Let $f_l(W^lx)$ and $f_n(W^nx)$ denote the two different models to be trained on loss function $L(f_l, f_n)$. We have explicitly written weight matrices $W^l$ and $W^n$ to emphasize that these two models start with a learnable linear layer. The following parametrization allows one to perform $L_1$ feature selection with both models:
\begin{align}
    \E_x[L(f_l(W^l (U\odot x)), f_n(W^n (U\odot x)))] \nonumber\\
    + \kappa(||W^l||_2^2 + ||W^n||_2^2 + ||U||_2^2),
\end{align}
where ${\rm dim}(U) = {\rm dim}(x)$, and $\E_x$ denotes averaging over the training set. Note that the input to the two models is masked by the same vector $U$: this is crucial; without $U$, we are just training an ensemble of independent models, whereas $U$ makes them coupled. Each $U_i$ is a redundant parameter, and this is equivalent to performing $L_1$ penalty on $W^l_{:i}$ and $W^n_{:i}$ together by Theorem~\ref{theo: main 2}. In the experiment, we let $f_l$ be a simple linear regressor without bias and $f_n$ be a three-layer feedforward network with the ReLU activation. For simplicity, we set the objective function $L(f_l, f_n) = CE(f_l(W^l(U\odot x), y) + CE(f_n(W^l(U\odot x), y))$ to be the summation of two Cross Entropy (CE) losses.

\begin{figure*}[t!]
    \centering
    \includegraphics[width=0.32\linewidth]{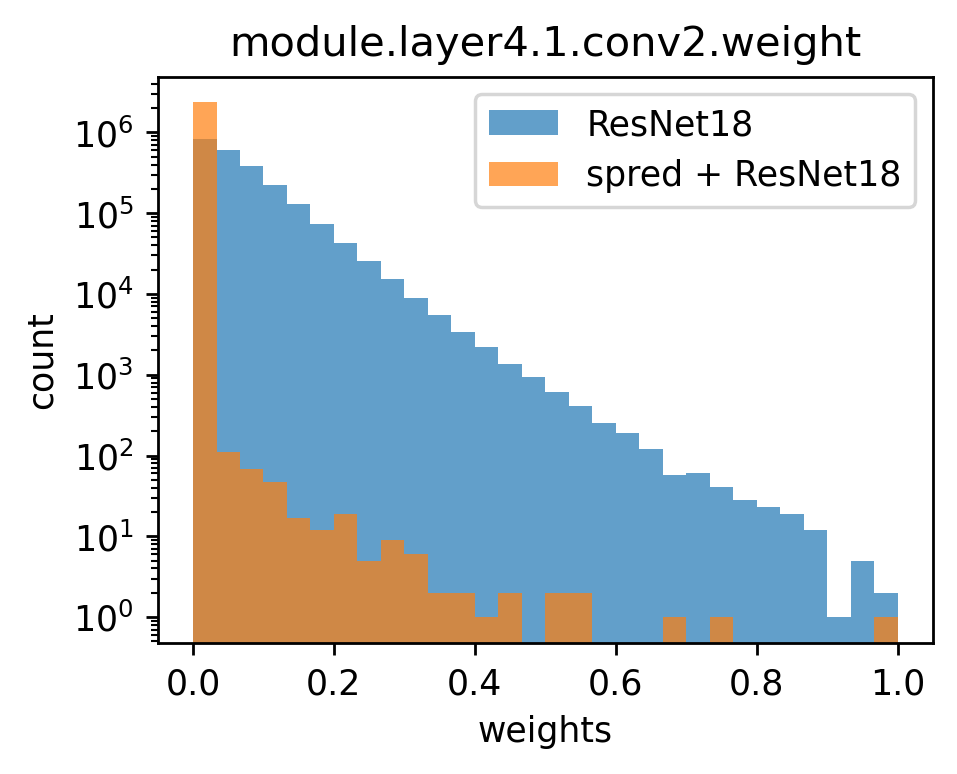}
    \includegraphics[width=0.32\linewidth]{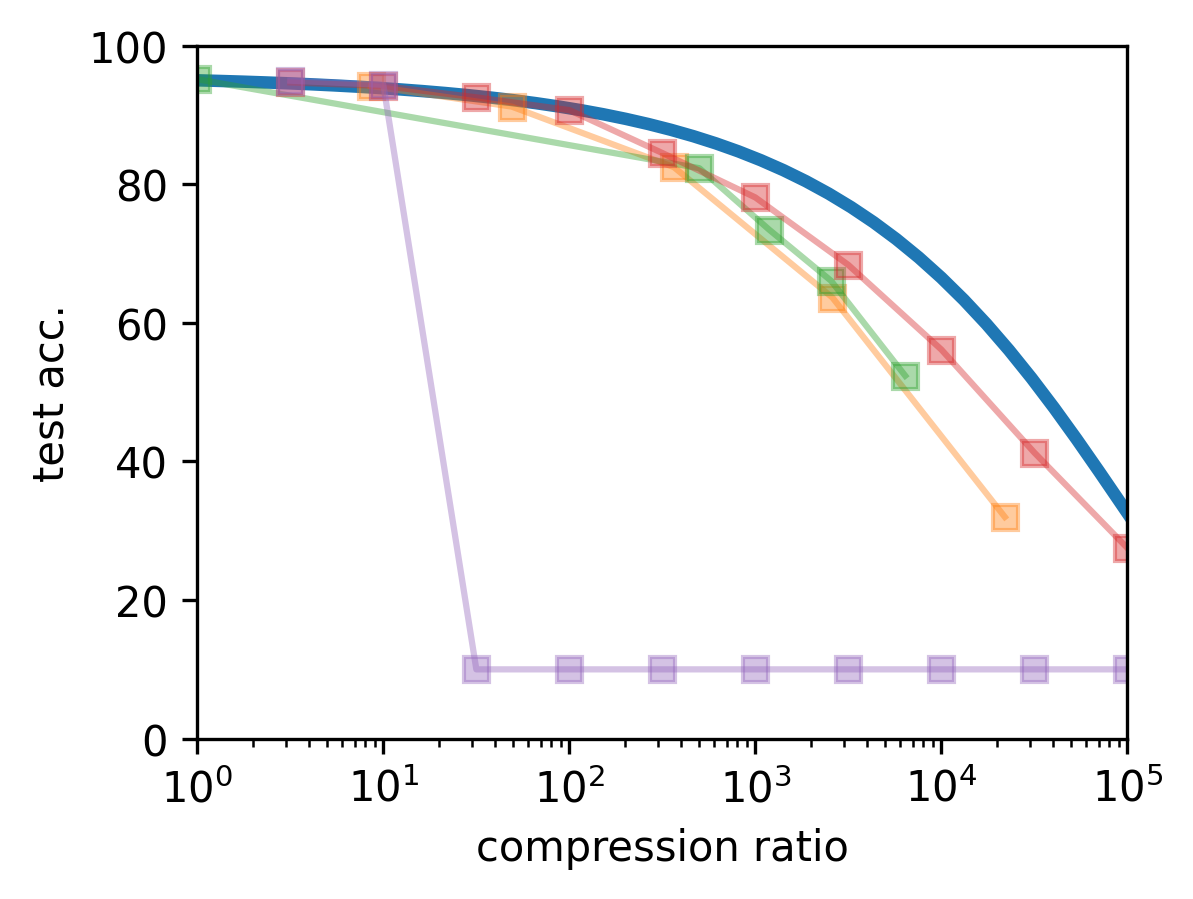}
    \includegraphics[width=0.32\linewidth]{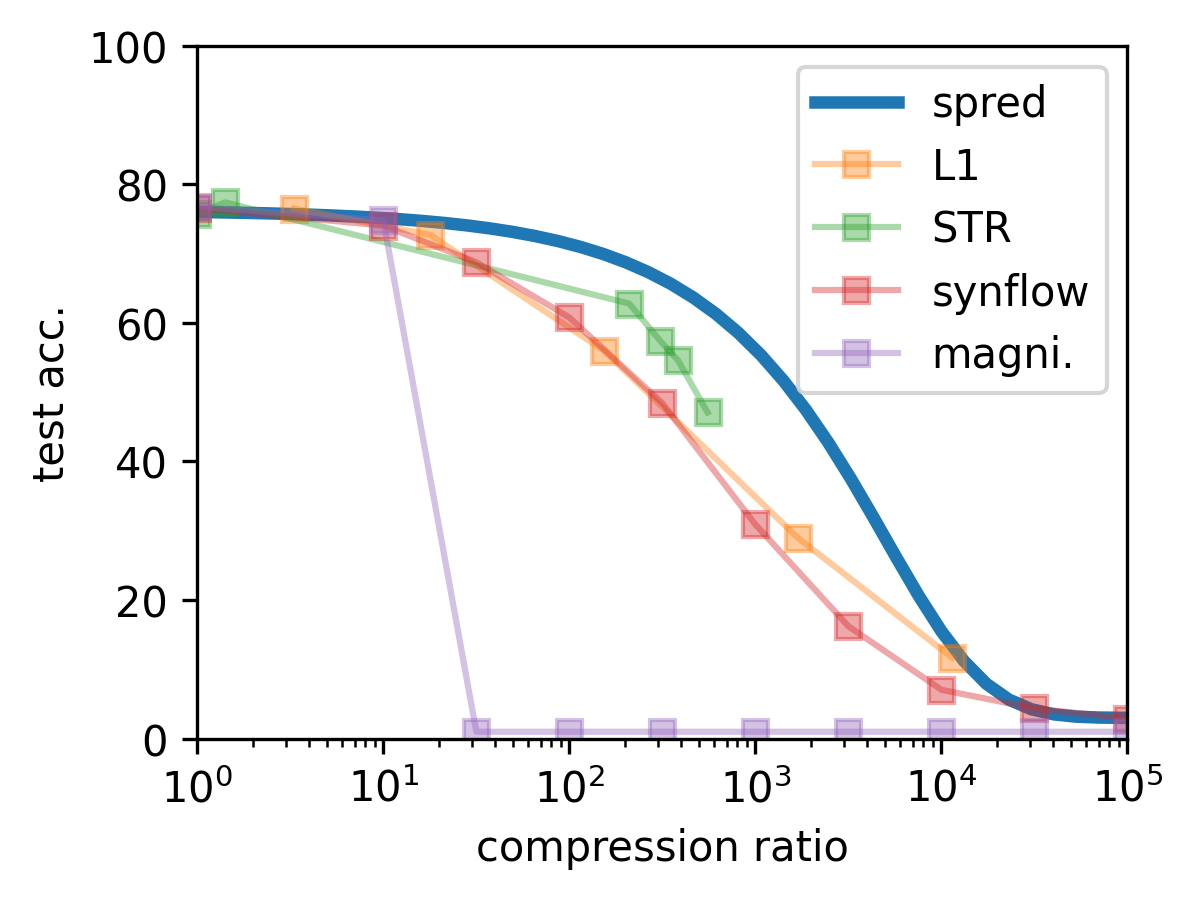}
    \vspace{-1em}
    \caption{\small Performance of $L_1$-based ResNet18 pruning on CIFAR-10 and CIFAR-100. For both datasets, the performance of \textit{spred} is competitive against any known existing pruning methods. \textbf{Left}: Distribution of weight parameters in the largest convolution layer of ResNet18 trained on CIFAR10. Training with \textit{spred} leads to a very sparse distribution without affecting test accuracy. \textbf{Mid}: CIFAR-10. \textbf{Right}: CIFAR-100.}
    \label{fig:resnet distribution main}
    \vspace{-1em}
\end{figure*}

See Table~\ref{tab: feature selection results} for the results. Because the dataset size is small, for each run of each model, we randomly pick 20\% samples as the test set, 20\% as the validation set for hyperparameter tuning, and 60\% as the training set. For SGD-based models (MLP, Linear, Linear + MLP), we stop the optimization when the accuracy on the validation set is not increasing. The performance is averaged over 20 independent samplings of the datasets for comparison. We report the percentage of the majority class of each dataset to justify whether the models produce meaningful results. In table~\ref{tab: feature selection results}, MLP contains one hidden layer of 4096 neurons. $f_n$ contains two hidden layers of 1024 neurons. \textit{spred} models are optimized by SGD. The learning rate and $\kappa$ are both selected from \{7e-1, 5e-1, 3e-1, 1e-1, 5e-2, 3e-2, 1e-2\}. Besides the deep learning methods, we also compare with HSIC-Lasso, a conventional $L_1$-based non-linear feature selection method~\citep{yamada2014high}, which has been a standard method, and recent works have identified it as one of the best-performing methods for these tasks \citep{sun2015feature, krakovska2019performance}.

We see that deep learning combined with \textit{spred} achieves state-of-the-art performance, outperformed by the conventional method on only one dataset. In sharp contrast, simply applying deep learning does not work on any of the datasets. This is expected for tasks whose dimension is far larger than the number of available data points because memorization can be too easy. Importantly, simply applying $L_1$ to an MLP fails badly because gradient descent cannot find a sparse solution and thus cannot prevent overfitting. In the future, designing better architectures that suit the task of gene selection will further boost performance.

\vspace{-0mm}
\subsection{Neural Network Compression}
\vspace{-0mm}

The proposed method also offers a principled way of performing network compression in deep learning. We experiment with unstructured weight sparsity for deep neural networks. Our method can also achieve structured compression, which we leave to future work.\footnote{For example, applying a vector of filter masks to the filters allows one to learn a sparse set of filters in CNN.} We apply \textit{spred} to all the weights of a ResNet in this section. Previous methods often rely on heuristics for pruning, such as removing the weights with the smallest magnitudes from a trained network. However, the problem with such methods is that one does not know, in principle, the effect of removing such weights, even though they seem to work empirically. Our method is equivalent to $L_1$ and has its theoretical foundation in both traditional statistics and Bayesian learning with a Laplace prior. The meaning of removing a parameter with magnitude $c$ is clear: its removal from the model will cause the training loss to increase by roughly $\kappa c$. We also emphasize that we are not proposing a new compression method: $L_1$ is known to lead to sparsity, and \textit{spred} is just a method for optimizing $L_1$ constraints. The performance of the proposed method can thus be no better than what a simple $L_1$ constraint can provide. The thesis of this section is that when an efficient way to optimize $L_1$ exists, it can perform as well as the existing methods that are not $L_1$-based, and thus $L_1$ based strategies are really worth exploring by the community. Prior to our work, many works have attempted to naively optimize the $L_1$ constraint with SGD, but such approaches have been found to perform rather badly compared with non $L_1$-based methods \citep{han2015deep}. 

We first train a ResNet18 on CIFAR10 with and without \textit{spred} both at $\kappa=5e-4$ and compare the weight distribution. Our implementation of ResNet18 contains roughly $11$M parameters, consistent with the standard implementation. See Figure~\ref{fig:resnet distribution main}. Both models achieve the established accuracy of $93\%$ while the training with \textit{spred} leads to a much sparser distribution. We now test the performance of $L_1$ for network pruning on CIFAR-10 and CIFAR-100. We implement \textit{spred} in the training protocol provided by \cite{kusupati2020soft}. We run the model at different weight decay strengths and report the pareto frontier obtained by fitting a sigmoid. For the raw data used to estimate the pareto frontier, see Appendix~\ref{app sec: exp}.

We compare with the following baselines. \textbf{L1 regularization}: this is the simplest baseline suggested by \citet{han2015deep} by simply adding an $L_1$ penalty to all the model parameters; we then prune at a given threshold and evaluate. The only hyperparameter is the regularization strength, which we search from $10^{-5}$ to $10^{-2}$. Soft threshold weight parametrization (\textbf{STR}): this is the $L_1$-based state-of-the-art neural network compression method. It serves as the main baseline of the proposed method because (1) it admits a direct interpretation as an $L_1$ approximate (but not exact), and (2) it uses a similar but nonequivalent reparametrization trick. Our results for L1 and STR are directly obtained using the implementation of \cite{kusupati2020soft}. Magnitude pruning (\textbf{magni.}): this is a simple method recommended by \cite{gale2019state} as a strong baseline that performs as well as the state-of-the-art methods in training a sparse network. \textbf{Synflow}: this method performs pruning at the beginning of training and is the state-of-the-art method for extreme compression rates. For example, with ResNet18 on CIFAR10/100, it is the only established benchmark that can prune beyond a $1000$ compression ratio \cite{tanaka2020pruning}. We use the implementation of \cite{tanaka2020pruning} to evaluate magnitude pruning and Synflow. For all baselines, we follow the hyperparameters recommended by \cite{kusupati2020soft} and \cite{tanaka2020pruning}, respectively. The comparison metric is the compression ratio, which is the total number of weights over the number of nonzero weights.

See the mid (CIFAR-10) and right panels (CIFAR-100). For both datasets, the training at $\kappa=5e-4$ recovers the standard performance of these models. For CIFAR10, the model can be pruned up to a 500 compression ratio while keeping an $>90\%$ accuracy. This is five times sparser than all the baselines for this performance level. For CIFAR100, the result is similar. The proposed method maintains a $>70\%$ accuracy while being an order of magnitude sparser than the previous methods.
 
To the extreme end, the proposed method keeps an above-chance accuracy even at a compression ratio of $10^5$ for both datasets, with a performance much better than Synflow, the best-known method at an extreme compression ratio. We also note that both STR and magnitude pruning has difficulty in extreme compression regime. The magnitude pruning method drops to chance-level accuracy at a compression ratio of 10, while STR cannot run above the 1000 compression ratio. Notably, our method has the implementation and training advantage over many existing methods in deep learning. One popular trick in network pruning is to iteratively retrain the model, having obtained a pruning mask, whereas the proposed method does not require iterative retraining.

\subsection{Compressing Pretrained Models}

\begin{figure}
    \centering
    \includegraphics[width=0.8\linewidth]{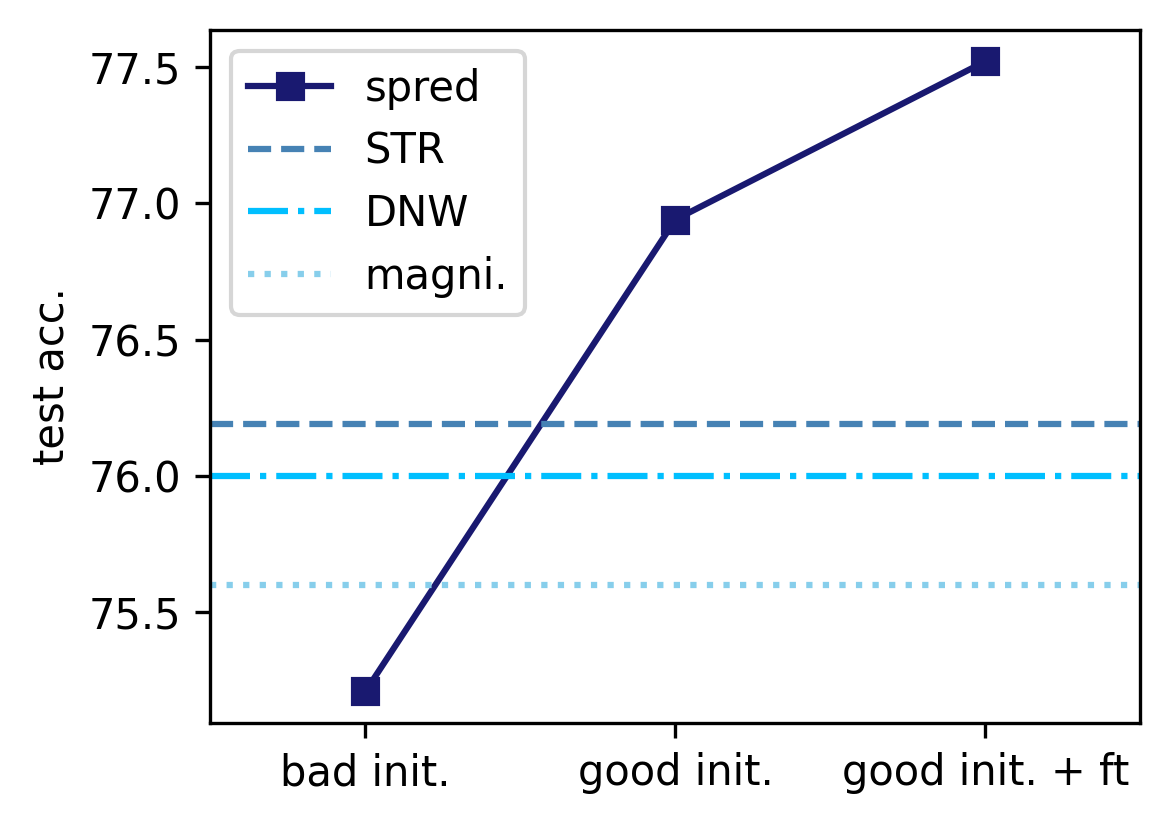}
    \vspace{-1em}
    \caption{\small \textit{spred} can be directly applied to standard pretrained models. The figure shows the top-1 test accuracy of ResNet50 on Imagenet at $80\%$ sparsity. Good and bad init. refers to initializing \textit{spred} with two different pretrained models, with $80\%$ and $77$ percent accuracy respectively. \textit{ft} refers to finetuning after pruning. We see that with a good pretrained model, \textit{spred} achieves the state-of-the-art result for the network compression task on Imagenet. Finetuning improves \textit{spred} further.}
    \label{fig:imagenet}
    \vspace{-1em}
\end{figure}

The training can be prohibitively expensive for larger tasks such as Imagenet and even larger tasks such as large language models. The important question is, therefore, whether we can perform efficient compression while leveraging the existing pretrained models. This experiment shows that \textit{spred} can be applied to existing pretrained models to achieve state-of-the-art network compression performance. For every pretrained weight matrix $\bar V_{ij}$, the weight $W_{ij}$ and the redundant weight $U_{ij}$ are initialized by 
\begin{align}
    W_{ij} = \sgn(\bar V_{ij})\sqrt{|\bar V_{ij}|}, \quad U_{ij} = \sqrt{|\bar V_{ij}|}.
\end{align}
The initialized model is then optimized by \textit{spred}. After the \textit{spred} training, the resulting weights $V_{ij} = U_{ij} W_{ij}$ are pruned towards a target sparsity to obtain the sparsed one $\hat V_{ij}$. The performance of \textit{spred} pruning weights $\hat V$ can be further improved by finetuning the model without \textit{spred} while setting the redundant weight as a boolean mask, where
\begin{align}
    W_{ij} = \hat{V}_{ij}, \quad U_{ij} = 1_{\hat V_{ij} > 0}.
\end{align}
The performances of \textit{spred} before and after finetuning are shown in Figure~\ref{fig:imagenet}. The \textit{spred} $\kappa$ is selected from $\{1e-5, 2e-5, 3e-5, 1e-4\}$ and $2e-5$ is found to be the most suitable for 80\% sparsity. We see that the performance of \textit{spred} with proper dense initialization outperforms previous baselines such as STR~\citep{kusupati2020soft}, magnitute~\citep{gale2019state}, and DNW~\citep{wortsman2019discovering} and achieves the state-of-the-art performance for Imagenet.

Our result thus promotes using the $L_1$ penalty in deep learning. Interestingly, the higher the $\kappa$, the more suited the trained model becomes for more aggressive pruning. $\kappa$ is thus a parameter worth finetuning to achieve the best sparsity-performance tradeoff. We also tried using the thresholds of the trained model as a mask, which we apply to a model at initialization, and a similar performance to the finetuned model is obtained. Our result thus supports the lottery ticket hypothesis and can be an alternative method for obtaining a lottery ticket. At a conceptual level, we have demonstrated this: $L_1$ can indeed work in the context of deep learning if we have an efficient way to optimize it.

\vspace{-0mm}
\subsection{Memory Cost}
\vspace{-0mm}

One might worry that using \textit{spred} will tend to double the memory cost of training. Our experiment shows that this is not true because the dominant factor of memory cost in training is minibatch size. At minibatch size $1$, \textit{spred} roughly doubles the memory cost of training. However, when the minibatch size is above $50$, the memory cost of \textit{spred} is of no observable difference from that of a standard network. See Appendix~\ref{app sec: memory cost}. For the same setting, we note that the time it takes for every training epoch is also only negligibly more than standard training by roughly $5\%$.

\vspace{-0mm}
\section{Discussion}
\vspace{-0mm}

In this work, we have thoroughly studied the landscape of a reparametrization trick that can be used to minimize a general nonconvex objective with an $L_1$ penalty. While the origin of the method itself is difficult to trace, we are the first to thoroughly investigate its theoretical influence on the loss landscape and to demonstrate how to apply it to deep learning. Our theory directly suggests that even in the case of highly complicated nonconvex landscapes, one may be able to optimize such a loss landscape highly efficiently with gradient descent. Our empirical result, in turn, demonstrates that $L_1$ penalty can help solve deep learning-related tasks very effectively. For all problems we approached, we have applied \textit{spred} in a straightforward way, and developing more sophisticated training methods for \textit{spred} is certainly one promising future direction.

\bibliographystyle{apalike}
\bibliography{ref}

\clearpage
\appendix
\onecolumn
\section{Experimental Concerns}\label{app sec: exp}

\subsection{Convergence of SGD on \textit{spred} Lasso}
\begin{figure}[t!]
    \centering
    \includegraphics[width=.35\linewidth]{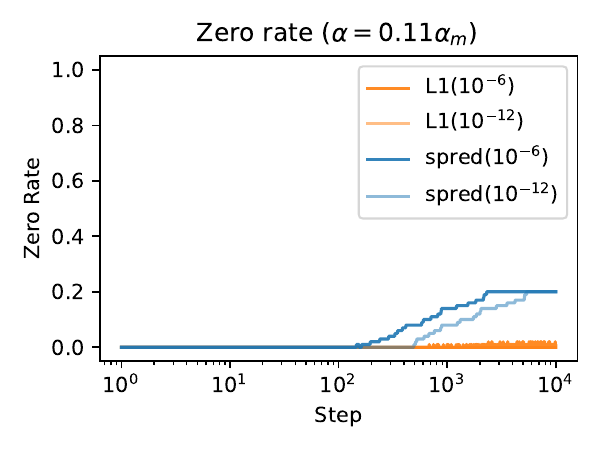}
    \includegraphics[width=.35\linewidth]{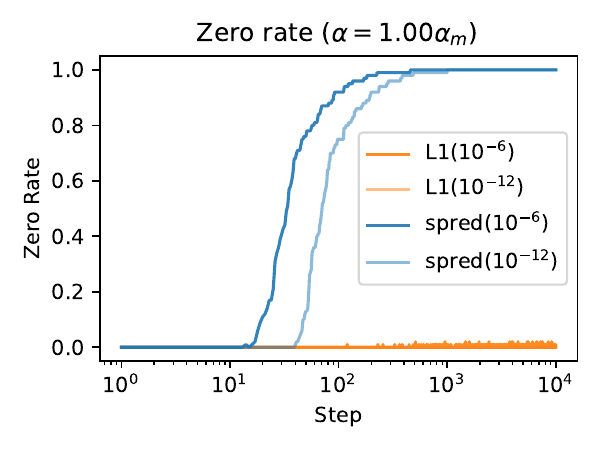}
    \vspace{-1em}
    \caption{The training trajectory of \textbf{L1} (GD on the vanilla lasso objective) and \textit{spred} when $\alpha\approx 0.3$ (left) and $\alpha\approx 2.3$(right)}
    \label{fig:L1-vs-rs-at-alphas}
\end{figure}

When using \textit{Spred}, one can speed up training is to set a threshold below which we set the parameter to zero at the stopping point. We test two levels of threshold, and both agree with the optimal solution at convergence. Figure~\ref{fig:L1-vs-rs-at-alphas} presents the training trajectory under different value of $\alpha$, and we see that they converge to the same value at roughly the same time scale. This can be used as a criterion for assessing the convergence of \textit{spred}.

\begin{figure}
    \centering
    \includegraphics[width=0.45\linewidth]{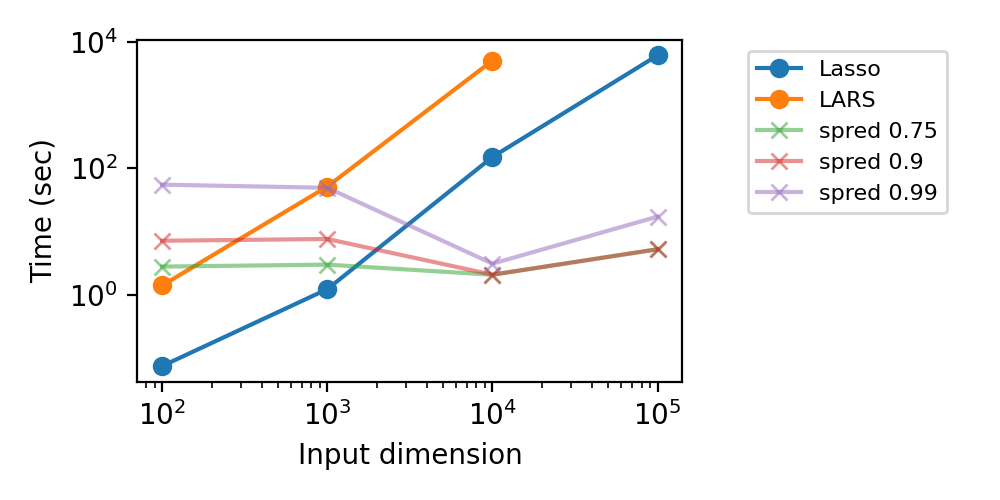}
    \includegraphics[width=0.45\linewidth]{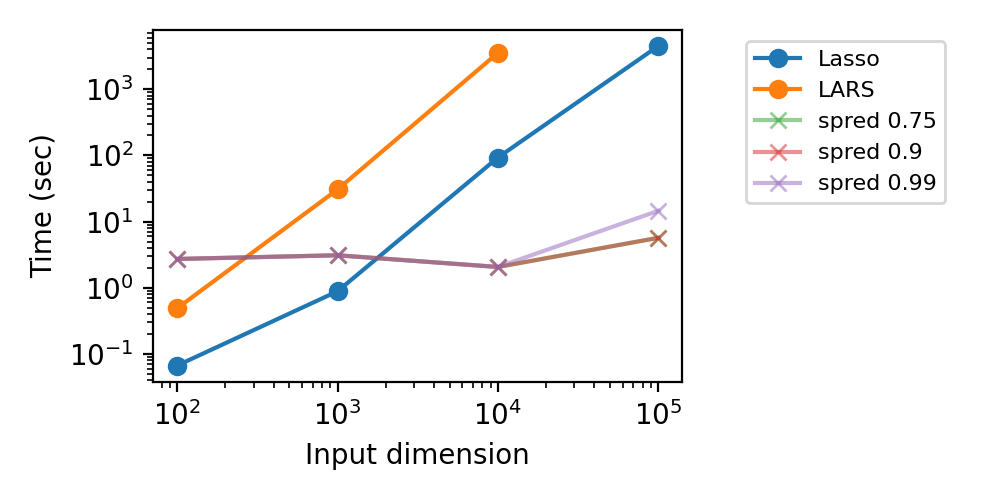}
    \includegraphics[width=0.45\linewidth]{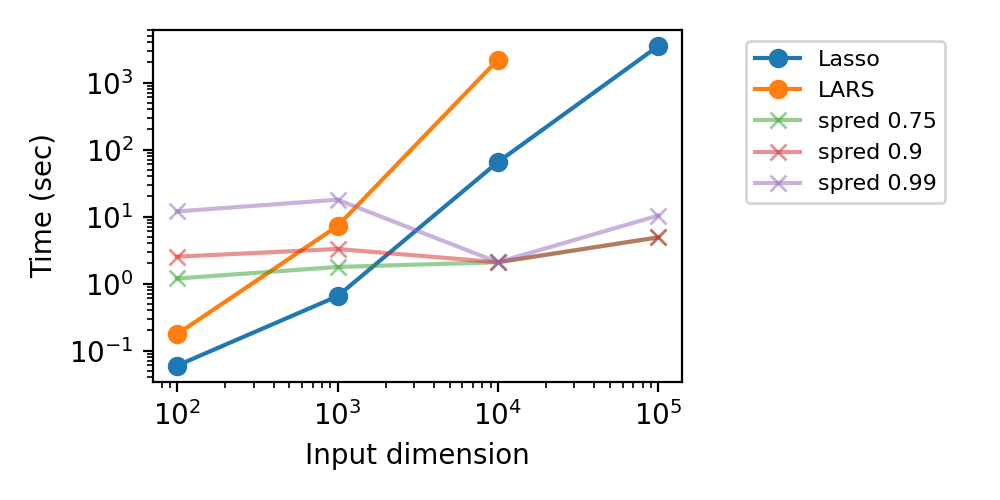}
    \includegraphics[width=0.45\linewidth]{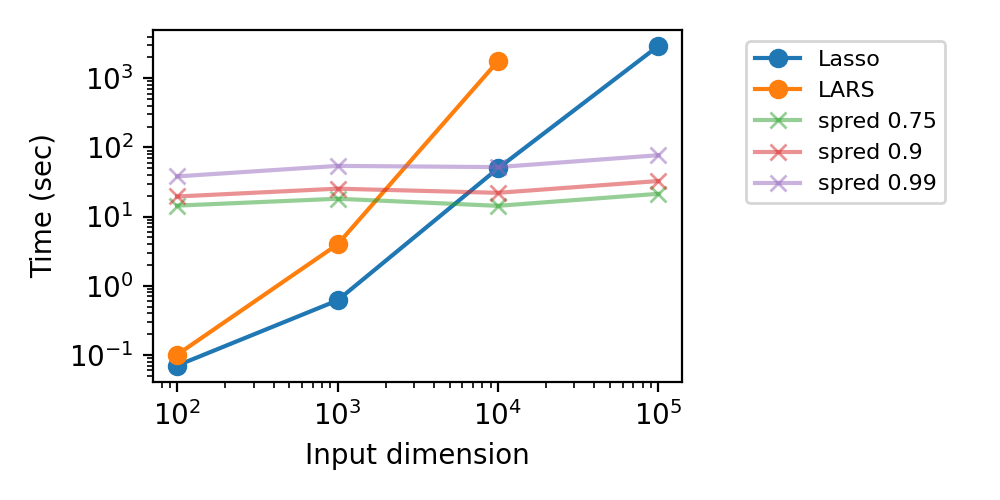}
    \vspace{-1em}
    \caption{{Performance of \textit{spred} across different values of $\alpha$. From upper left to lower right: $\alpha =2,\ 3,\ 4,\ 5$. The main text contains the case when $\alpha=1$.}}
    \label{app fig:lasso different alpha}
\end{figure}

\subsection{\textit{spred} linear regression for different regularization strengths}

Now, we compare the optimization efficiency of \textit{spred} with the coordinate descent and LARS solutions of lasso under different input and output dimensions. The coordinated descent solution of lasso is denoted as \textit{Lasso}. The Least Angle Regression of lasso is denoted by \textit{LARS}. See Figure~\ref{app fig:lasso different alpha}. For \textit{spred}, we report the time when the zero rates of the solution matrix hit $75\%$, $90\%$, and $100\%$ of the zero rates of the converged solution. We note that there is no discernible difference in the training loss with the lasso objective for all three rates. As the plots show, the optimization speed of \textit{spred} compares rather favorably against the standard methods at a large data dimension.

\clearpage

\begin{figure}[t!]
    \centering
    \includegraphics[width=0.32\linewidth]{plots/resnet3.png}
    \includegraphics[width=0.32\linewidth]{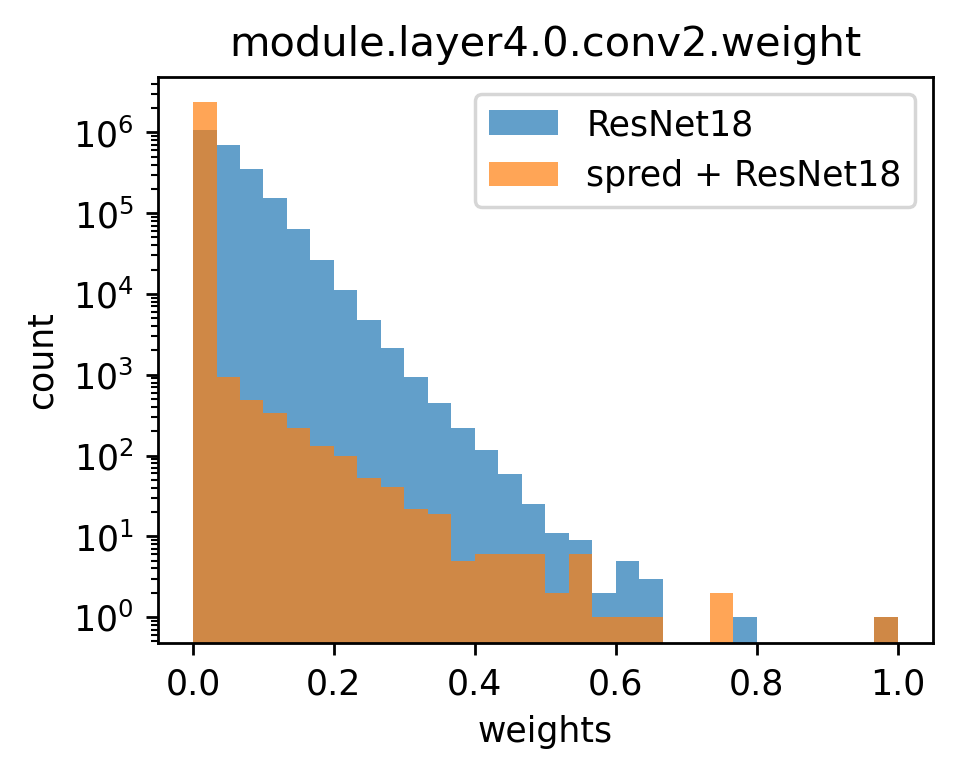}
    \includegraphics[width=0.32\linewidth]{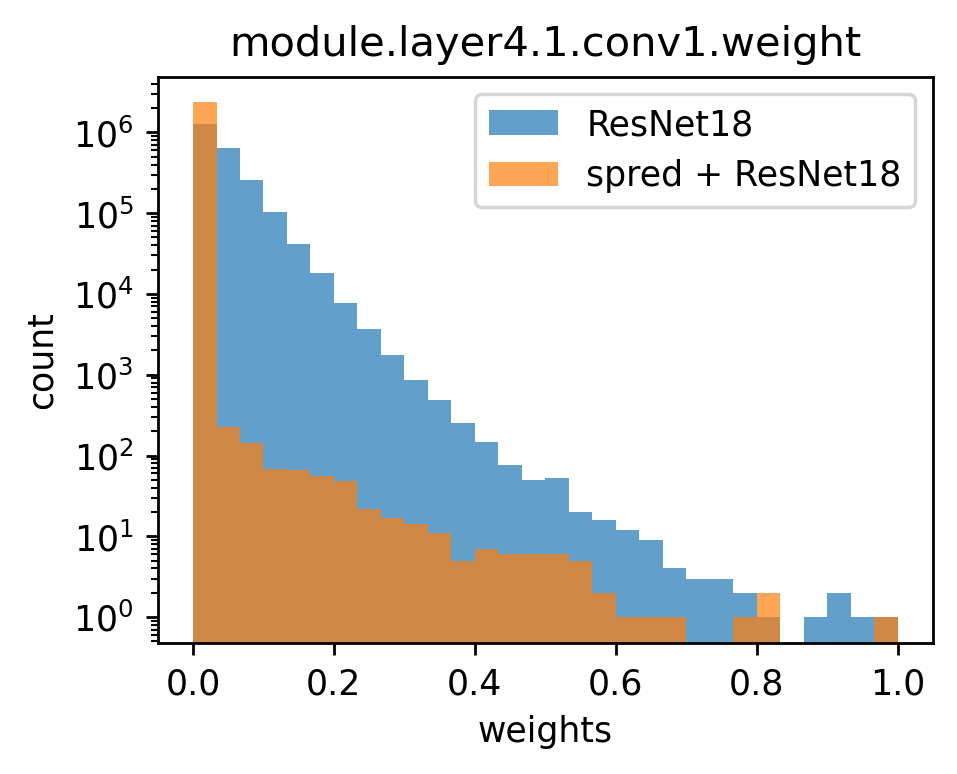}
    \caption{Normalized parameter distribution of the three largest convolutional layers of ResNet18 trained on CIFAR10 with SGD. The blue histogram shows the distribution of a normal ResNet18 with weight decay strength $5\times 10^{-4}$, which is very dense. The orange shows the distribution of a \textit{spred} ResNet18 (also with $5\times 10^{-4}$ weight decay), which exhibits a predominant peak at zero that includes more than $99.9\%$ of all the weight parameters of the layer. This shows that training with a very small value of regularization with \textit{spred} already leads to a parameter distribution that favors sparsity.}
    \label{fig:resnet distribution}
\end{figure}

\subsection{Weight Distribution of a Trained ResNet}
We show more results on the weight distribution of a trained ResNet18, with roughly $11$M parameters in total. We plot the parameter distribution of the three largest convolutional layers, each with roughly $2.3$M parameters. See Figure~\ref{fig:resnet distribution}.

\subsection{Detailed Description of the Feature Selection Task}
See Table~\ref{tab: cancer dataset statistics} for the statistics of the datasets. The datasets are taken from the public datasets of Gene Expression Omnibus.\footnote{\url{https://www.ncbi.nlm.nih.gov/geo/}} The indices of the datasets are the same as the indices on GEO.

\begin{table}[t!]
\centering
\caption{Basic statistics of seven gene datasets.}\label{tab: cancer dataset statistics}
\begin{tabular}{lllll}
\hline
Dataset & \#features & \#labels & \#samples & $\frac{\#\text{samples}}{\#\text{features}}$ \\\hline
GDS1815 \citep{phillips2006molecular}      & 22283      & 15       & 400       & 1.79\%               \\
GDS1816 \citep{phillips2006molecular}      & 22645      & 15       & 400       & 1.77\%               \\
GDS3268 \citep{noble2008regional}      & 44290      & 8        & 606       & 1.37\%               \\
GDS3952 \citep{labreche2011integrating}      & 54675      & 8        & 324       & 0.59\%               \\
GDS4761 \citep{kimbung2014claudin}      & 52378      & 7        & 91        & 0.17\%               \\
GDS5027 \citep{prat2014based}      & 54675      & 6        & 468       & 0.86\%              \\\hline
\end{tabular}
\end{table}



\clearpage

\vspace{-1em}
\subsection{Estimating the Pareto Frontier for Network Compression}
\begin{figure}[t!]
    \centering
    \includegraphics[width=0.45\linewidth]{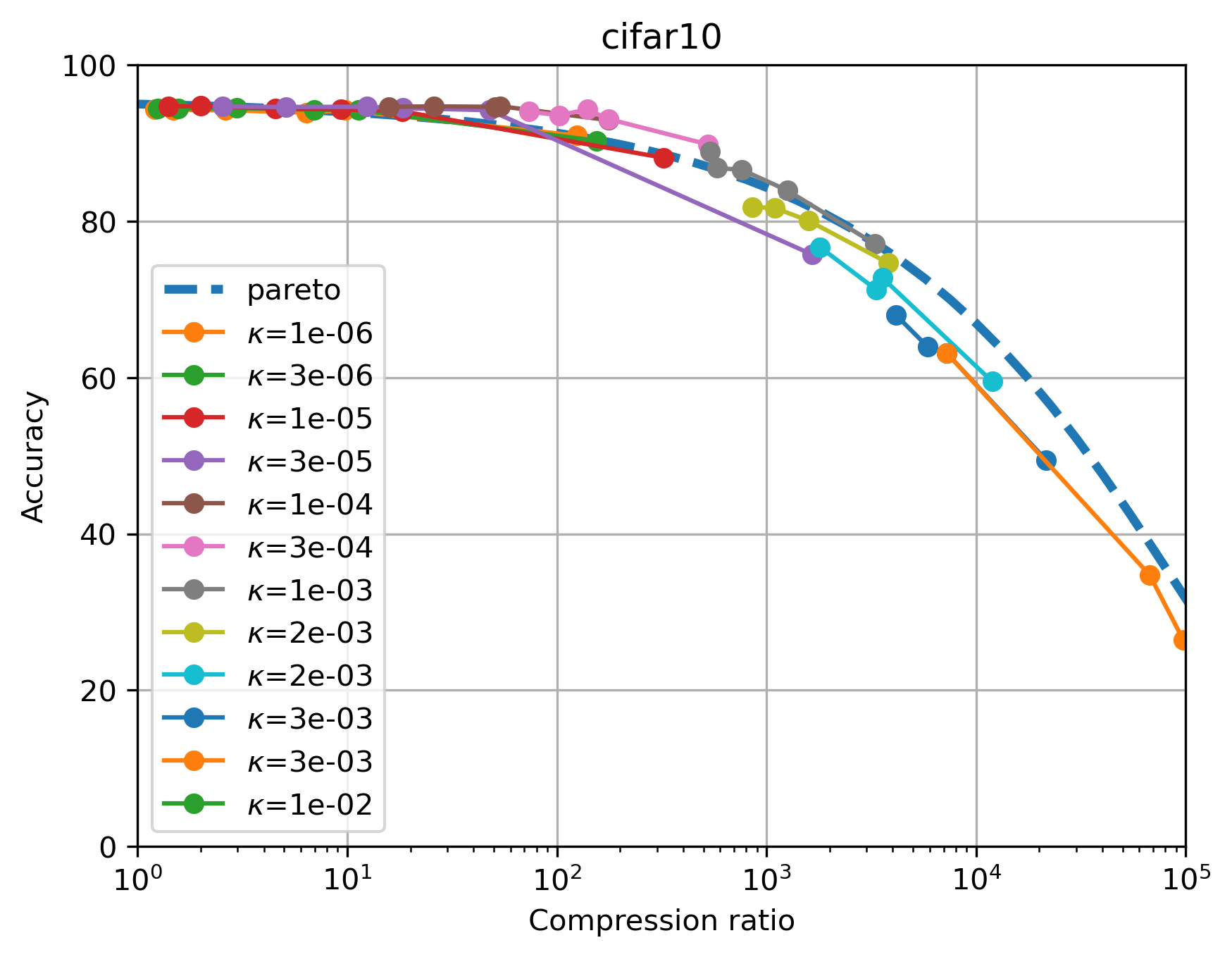}
    \includegraphics[width=0.45\linewidth]{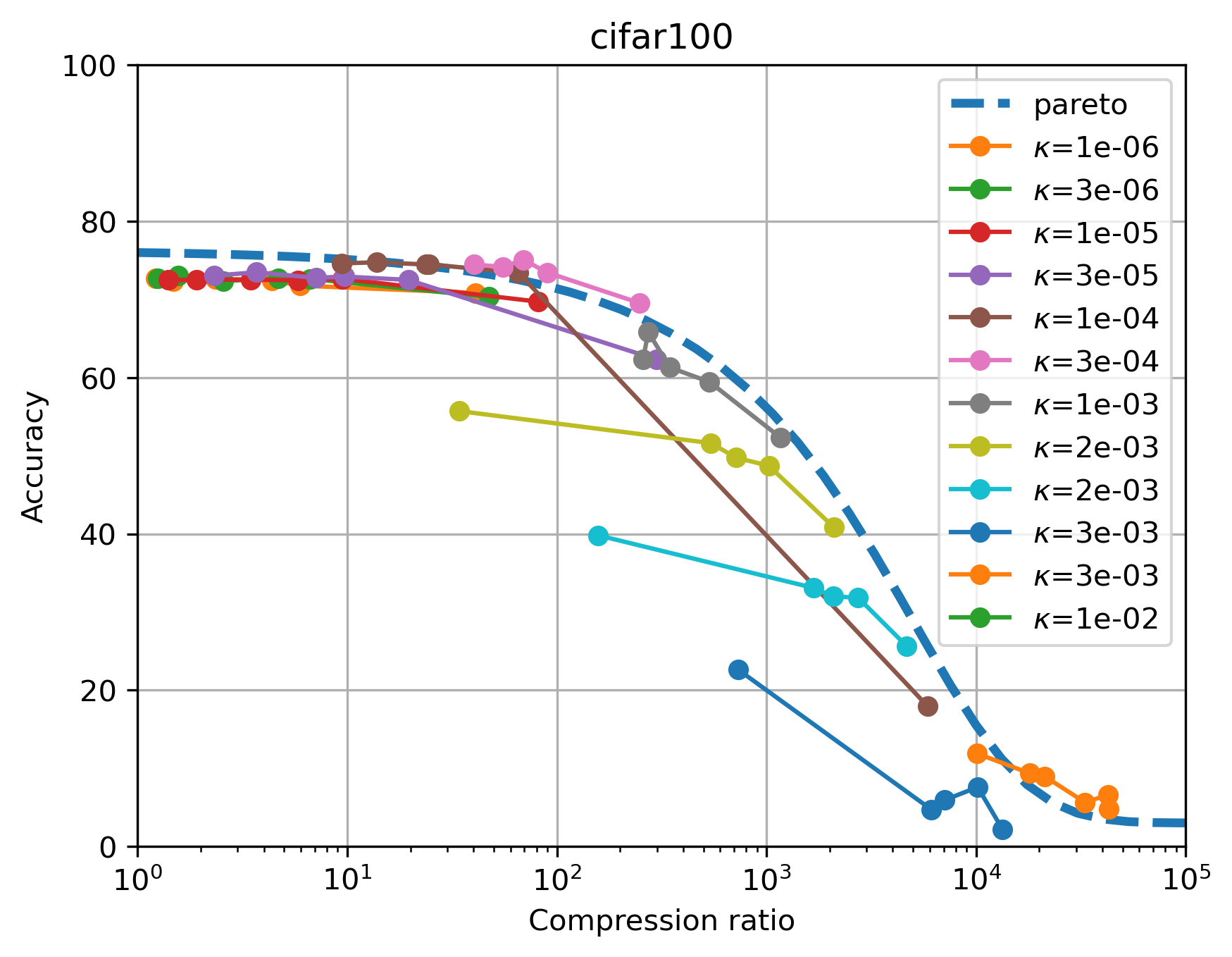}
    
    \caption{Pareto frontier of \textit{spred} for CIFAR10 and CIFAR100.}
    \label{fig:pareto raw data}
\end{figure}
See Figure~\ref{fig:pareto raw data}. The grey dashed line shows the estimated pareto frontier with a tanh function.

\subsection{Memory Cost}\label{app sec: memory cost}
\begin{figure*}
    \centering
    \includegraphics[width=.8\linewidth]{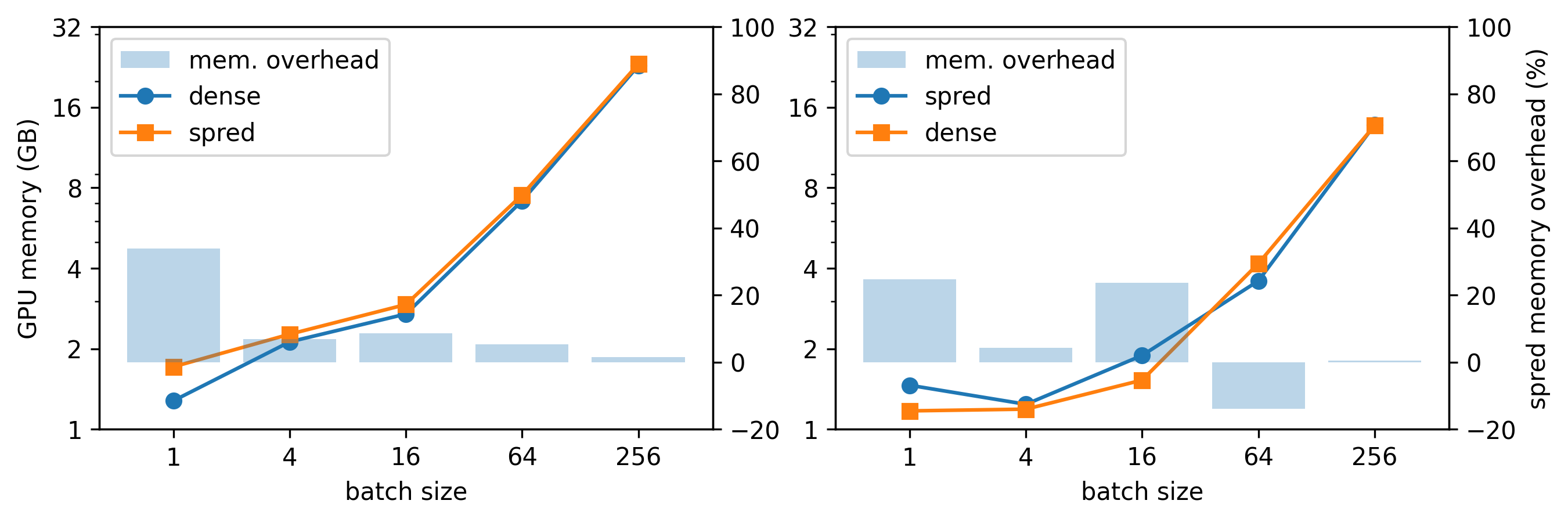}
    \caption{The memory cost of \textit{spred} is negligible at common batch sizes for Imagenet. Training with batch size $1$, \textit{spred} doubles the memory cost of training. However, the relative cost overhead diminishes to zero when the batch size is of order $10^2$. \textbf{Left}: ResNet50. Right: MobileNetV1.}
    \label{fig:memory cost}
\end{figure*}

In this section, we compare the memory cost of \textit{spred} with standard training on two different architectures. See Figure~\ref{fig:memory cost}.

\clearpage
\section{Proof}\label{app sec: proofs}
\subsection{Proof of Theorem~\ref{theo: sparsity of parameters}}
For notational conciseness, we prove the case when $\alpha =\beta = \kappa$. The case $\alpha \neq \beta$ can be reduced to this simpler case if we redefine both $U$ and $W$ by a constant scaling. We first prove a lemma. 
\begin{lemma}\label{lemma 1}
    For all $i$, any local minimum of Eq.~\eqref{eq: rs loss general} satisfies 
    \begin{equation}
        |U_i| = |W_i|.
    \end{equation}
\end{lemma}
\textit{Proof}. We prove by contradiction. Suppose not. Then there exists $U'$, $W'$ and index $i$ such that $|U_i| \neq |W_i|$ and they are a local minimum of $L(U' \odot W') + L_2\ reg$, where
\begin{equation}
    L_2\ reg. = \kappa(||U||_2^2 + ||W||_2^2).
\end{equation}
Now, we consider an infinitesimal perturbation of the solution such that $U_i = U_i'(1 + dz)$ and $W_i = W_i'(1-dz)$. It is straightforward to see that, by the definition of element-wise multiplication,
\begin{equation}
    L(U' \odot W') = L(U \odot W).
\end{equation}
Without loss of generality, we assume $|U_i| < |W_i|$. Now, because $U_i < W_i$, the $L_2\ reg.$ term strictly reduces:
\begin{gather}
    U_i'^2(1+2dz) + W_i'^2 (1 -2dz) - U_i^2 - W_i^2 = 2(U_i'^2 - W_i'^2)dz <0.
\end{gather}
This means that $U_i$ and $W_i$ cannot be a local minimum. The proof is complete. $\square$

The above lemma implies that to find the global minimum of Eq.~\eqref{eq: rs loss general}, it suffices to minimize over the solutions such that $|W_i| = |U_i|$ for all $i$. The following lemma shows that the two loss function are identical if we restrict to the domain where $|W_i| = |U_i|$.

\begin{lemma}\label{lemma 2}
    Let $W \odot U =V$ and $|W_i| = |U_i|$ for all $i$. Then,
    \begin{equation}
        L_{rs}(W, U) = L_{L1}(V).
    \end{equation}
\end{lemma}

\textit{Proof}. When $|W_i| = |U_i|$,
\begin{align}
    L_{rs} &=  L(U \odot W) + \kappa\left(\sum_i U_i^2 + W_i^2 \right)\\
    &= L(U \odot W) + \kappa\left(\sum_i 2|U_i W_i| \right)\\
    &= L(U \odot W) + 2\kappa ||U\odot W||_1.
\end{align}
By definition, $U\odot W = V$, and so this loss is, in turn, equivalent to the following loss:
\begin{equation}
    L(V) + 2 \kappa ||V||_1.
\end{equation}
This finishes the proof. $\square$

Now, we are ready to prove the main theorem. To repeat, the main theorem states the following (when $\alpha=\beta$).

\begin{theorem}
    Let $\alpha\beta = \kappa^2$ and
    \begin{equation}
        L_{sr}(U, W, V_d) := L(U\odot W, V_d) + \alpha ||U||^2 + \beta ||W||^2.
    \end{equation}
    Then, $(U,W, V_d)$ is a global minimum of Eq.~\eqref{eq: rs loss general} if and only if (a) $|U_i| = |W_i|$ for all $i$ and (b) $(U\odot W, V_d)$ is a global minimum of Eq.~\eqref{eq: l1 loss general}.
\end{theorem}

\textit{Proof}. The theorem immediately follows from the combination of the previous two lemmas. $\square$

\subsection{Proof of Theorem~\ref{theo: benign landscape}}
To repeat, the theorem states the following.
\begin{theorem}
    All stationary points of Eq.~\eqref{eq: rs loss general} satisfy $|U_i| = |W_i|$. Additionally, $(U,W)$ is a local minimum of Eq.~\eqref{eq: rs loss general} if and only if (a) $V=U\odot W$ is a local minimum of Eq.~\eqref{eq: l1 loss general} and (b) $|U_i| = |W_i|$.
\end{theorem}
\textit{Proof}. We first prove the statement regarding the local minima. For both directions, we prove by contradiction. The forward direction is much easier to prove. Let $(U,W)$ be a local minimum of $L_{sr}$ and suppose $V$ is not a local minimum of $L_{L1}(V)$. Then, one can infinitesimally perturb $V$ such that $V + dz$ has a smaller loss. This corresponds to a perturbation in $U$ and $W$ under the constraint $|U_i| = |W_i|$. By Lemma~\ref{lemma 2}, $L_{rs}$ under this perturbation is also smaller than the unperturbed value. Thus, $(U,W)$ is not a local minimum -- a contradiction.

We now consider the backward direction.  Let $V$ be a local minimum of $L_{L1}$ and suppose $(U,W)$ is not a local minimum of $L_{rs}$. As Lemma~\ref{lemma 2} shows, if we restrict to the subspace where $|U_i| = |W_i|$, there cannot be a perturbation that leads to a lower loss value because in this subspace, $L_{rs}$ is equivalent to $L_{L1}$. Thus, that $(U,W)$ is not a local minimum implies that there exists perturbation $dz_U$ and $dz_W$ such that $(U + dz_U, W+dz_W)$ has a smaller loss value than $(U,W)$. The loss function value is 
\begin{equation}
    L((U+dz_U)\odot (W+dz_W)) + \kappa(||U+dz_U||_2^2 + ||W+dz_W||^2) < L_{rs}(U\odot W)
\end{equation}
such that $|(W+dz_W)_i| \neq |(U+dz_U)_i|$. Now, we can construct a new parameter $U' = \sgn(U+dz_U) \sqrt{|(U+dz_U)\odot (W+dz_W)|}$, $W'=\sgn(W+dz_W) \sqrt{|(U+dz_U)\odot (W+dz_W)|}$. This transformation is also infinitesimal and leaves the $L$ term unchanged. However, it strictly decreases the $L_2$ term
\begin{equation}
    \|U'\|^2 + \|W'\|^2 = 2|(U+dz_U)\odot (W+dz_W)|^2 < ||U+dz_U||_2^2 + ||W+dz_W||^2.
\end{equation}
Thus, we have constructed a model such that $|U'_i| = |W_i'|$ for all $i$, and with a strictly smaller loss. By Lemma~\ref{lemma 2}, this implies that $V$ is not a local minimum of $L_{L1}$. This is a contradiction.

Now we prove the statement regarding all the stationary points. Since we have proved Lemma~\ref{lemma 1}, it is sufficient to only prove the condition for all saddles points. We show that when $|U_i|\neq |W_i|$ the variation of Eq.~\eqref{eq: rs loss general} has a nonvanishing first-order variation if one varies $U$ and $W$ by a perturbative amount. Consider the following transformation of $U_i$ and $W_i$
\begin{equation}
\begin{cases}
    U_i \to U_i + dz,\\
    W_i \to W_i - dz.
\end{cases}
\end{equation}
To first order in $dz$, $L$ remains unchanged, whereas the regularization term changes by
\begin{equation}
     2 \kappa dz (U_i - W_i).
\end{equation}
Because $|U_i|\neq |W_i|$, this is a first order term in $dz$ and so $U_i$, $W_i$ cannot be a saddle. The proof is complete. $\square$

\subsection{Proof of Theorem~\ref{theo: benign saddle}}
To repeat, the theorem statement is the following.
\begin{theorem}
    Let $|U|= |W|$, $V=U\odot W$ and $L$ be everywhere differentiable. Then, for every infinitesimal variation $\delta V$,
    \begin{enumerate}
        \item if $L_{L1}(V)$ is directionally differentiable in $\delta V$, there exist variations $\delta W, \delta U\in \Theta(\delta V )$ such that $ L_{L1}(V+ \delta V)=  L_{rs}(U + \delta U, W+ \delta W)$;
        \item if $L_{L1}(V)$ is not directionally differentiable in $\delta V$, there exist variations $\delta W, \delta U\in \Theta\left((\delta V)^{0.5} \right)$ such that $ L_{L1}(V+ \delta V)=  L_{rs}(U + \delta U, W+ \delta W)$.
    \end{enumerate}
\end{theorem}
\textit{Proof}. Because $L_{L1}(V)= L_{rs}(U,W)$ when $|U|=|W|$, we have $L_{L1}(V +\delta V)= L_{rs}(U+ \delta U,W + \delta W)$ as long as
\begin{equation}\label{eq: variation condition}
    W \odot \delta U  + U \odot \delta W + \delta U \odot \delta W = \delta V,
\end{equation}
provided that the constraint $|U+\delta U| = |W + \delta W|$ is satisfied. Let $K$ denote the set indices such that for all $i\in K$, $V_i=0$. Because $L$ is differentiable, $L_1$ is directionally differentiable as long as $\delta V_i=0$ for all $i\in K$. This means that for all $i \notin K$, $|U_i| = |W_i|=\sqrt{|V|} \neq 0$. In turn, this means that with an infinitesimal $\delta V$, setting $\delta U_i = \delta V_i / 2W_i \in \Theta (\delta V)$ and $\delta W_i = \delta V_i / 2U_i \in \Theta (\delta V)$ achieves the desired variation:
\begin{equation}
    W \odot \delta U  + U \odot \delta W = \delta V.
\end{equation}
One can easily check that the constaint is also satisfied. This proves the first part of the theorem.

For the second part, we first note that $L_{L1}$ is only directionally nondifferentiable in $\delta V$ if for some $i\in K$, $\delta V_i \neq 0$. Since $V_i=0$, we have $U_i=W_i=0$, and so for these indices Eq.~\eqref{eq: variation condition} becomes
\begin{equation}
    \delta U_i \delta W_i = \delta V_i.
\end{equation}
Because the variation must also satisfy $|\delta U_i| = \delta |W_i|$, one solution is 
\begin{equation}
\begin{cases}
    \delta U_i = \sqrt{|\delta V_i|};\\
    \delta W_i = \sgn(V_i) \sqrt{|\delta V_i|}.
\end{cases}
\end{equation}
For infinitesimal $\delta V$, $\Theta(\delta V^{0.5} + \delta V) = \Theta(\delta V^{0.5})$. We thus have that $\delta U$, $\delta W \in \Theta(\delta V^{0.5})$. This proves the second part of the theorem. $\square$

\subsection{Proof of Theorem~\ref{theo: main 2}}
\begin{theorem}
    Let $\alpha\beta = \kappa^2$ and
    \begin{equation}
        L_{sr}(u, W, V_d) := L(uW, V_d) + \alpha u^2 + \beta ||W||^2.
    \end{equation}
    Then, $(u,W, V_d)$ is a global minimum of Eq.~\eqref{eq: rs loss general} if and only if (a) $|u| = ||W||_2$ for all $i$ and (b) $(u W, V_d)$ is a global minimum of Eq.~\eqref{eq: l1 loss general}.
\end{theorem}
The proof is similar to that of Theorem~\ref{theo: sparsity of parameters}, and we thus only give a proof sketch.

\textit{Proof Sketch}. When $|u| = ||W||_2$, it is easy to check that the two loss functions agree in value.  When $|u| \neq ||W||_2$, one can always find  continuous transformation (rescaling $u$ and $W$ simultaneously) of $u$ and $W$ such that the loss function is strictly reduced, and these points cannot be local minima.
$\square$

The proof also shows that every minimum of $L_{rs}$ corresponds to the local minimum in the original loss, consistent with Theorem~\ref{theo: benign landscape}.
This result can be immediately generalized to the case of multi-group $L_1$, where we want to apply $L_1$ (possibly with different strengths) to different groups. This can be proved by simply induction on the size of the set of groups and using Theorem~\ref{theo: main 2}.

\end{document}